# Segment Any Architectural Facades (SAAF)

## An automatic segmentation model for building facades, walls and windows based on multimodal semantics guidance

*Peilin Li[1], Jun Yin[2], Jing Zhong[3], Ran Luo[4], Pengyu Zeng[5], Miao Zhang[6]*
*[1]National University of Singapore, [2]Guangzhou University*
*[2,3,5,6]Tsinghua University, [4]South China University of Technology*
*[1]e1351227@u.nus.edu [4]201930093463@mail.scut.edu.cn*
*[2,3,5]{yinj24|zhongj24|zeng-py24}@mails.tsinghua.edu.cn*
*[6]zhangmiao@sz.tsinghua.edu.cn*

*In the context of the digital development of architecture, the automatic segmentation of walls and windows is a key step in improving the efficiency of building information models and computer-aided design. This study proposes an automatic segmentation model for building facade walls and windows based on multimodal semantic guidance, called Segment Any Architectural Facades (SAAF). First, SAAF has a multimodal semantic collaborative feature extraction mechanism. By combining natural language processing technology, it can fuse the semantic information in text descriptions with image features, enhancing the semantic understanding of building facade components. Second, we developed an end-to-end training framework that enables the model to autonomously learn the mapping relationship from text descriptions to image segmentation, reducing the influence of manual intervention on the segmentation results and improving the automation and robustness of the model. Finally, we conducted extensive experiments on multiple facade datasets. The segmentation results of SAAF outperformed existing methods in the mIoU metric, indicating that the SAAF model can maintain high-precision segmentation ability when faced with diverse datasets. Our model has made certain progress in improving the accuracy and generalization ability of the wall and window segmentation task. It is expected to provide a reference for the development of architectural computer vision technology and also explore new ideas and technical paths for the application of multimodal learning in the architectural field.*

## INTRODUCTION

The rapid progress in machine learning has stimulated extensive research across various domains, such as sustainable architectural practices (Zou et al., 2021; Zeng et al., 2025; Jia et al., n.d.), energy-efficient design strategies (Zhang et al., 2024; Zeng et al., 2025; Zeng et al., 2025b), multimodal content generation (He et al., 2024; Zhang et al., 2025; Wang et al., 2025; Sun et al., 2025a, 2025b, 2025c; Wang et al., 2025b; Gao et al., 2025), visual feature enhancement (Zhang et al., 2025; Wang et al., 2025; He et al., 2025; Yin et al., 2025), and collaborative human-AI systems (He et al., 2024; Zeng et al., 2024; Wang et al., 2025; Yin et al., 2024; Wang et al., 2025b). This technological momentum has also extended into architecture, particularly in efforts to digitize and automate the interpretation of built environments. Within the domain of architectural informatization, the automatic detection and interpretation of building components has emerged as a key focus at the convergence of architecture and computer vision. The accelerating advancement of deep learning and computer vision techniques has driven a growing body of research employing deep learning - based approaches—such as image segmentation, point cloud analysis, and semantic parsing—to extract component-level information from 2D architectural imagery (Hou et al., 2021).

Wall-to-Window Ratio (WWR) impacts building performance, aesthetics, and adaptability. Accurate wall-window segmentation is crucial for simulation, restoration, and modeling, providing precise geometric and semantic inputs for BIM and enhancing model automation.

However, facade complexity, lighting conditions, and decorative details reduce segmentation accuracy (Liu et al., 2020). Current methods face limitations: Manual methods are inefficient and subjective, while automated models generalize poorly. Tools like SAM (Kirillov et al., 2023) still rely on user input and lack full automation. Moreover, the limitations of unimodal image recognition systems further underscore the potential of this research. These limitations include insufficient annotations, ambiguous regional behaviors, complex material reflectance, and inadequate understanding of architectural semantics. The emergence of multimodal models, capable of integrating and correlating data from heterogeneous sources such as images, text, and point clouds (Baltrušaitis et al., 2018), has illuminated a feasible pathway to improve segmentation accuracy and applicability. However, their reliance on large datasets and higher training costs limits practical use in architecture.

To tackle these challenges, we propose Segment Any Architectural Facades (SAAF), a multimodal-guided automatic wall-window segmentation model (Figure 1) with three key advantages: (1) SAAF not only streamlines facade segmentation and analysis workflows but also empowers users to iteratively refine results via natural language descriptions. (2) To manage task complexity, we decompose it into subtasks and apply a modality decomposition mechanism, using vectorized data to link text and visuals while reducing multimodal training costs. (3) we design a deep learning-based network architecture to accurately identify and segment key elements in facade images. Extensive experiments suggest that SAAF achieves competitive mIoU performance compared to existing methods across diverse façade styles, indicating its potential as a useful tool for architectural design and workflow automation.

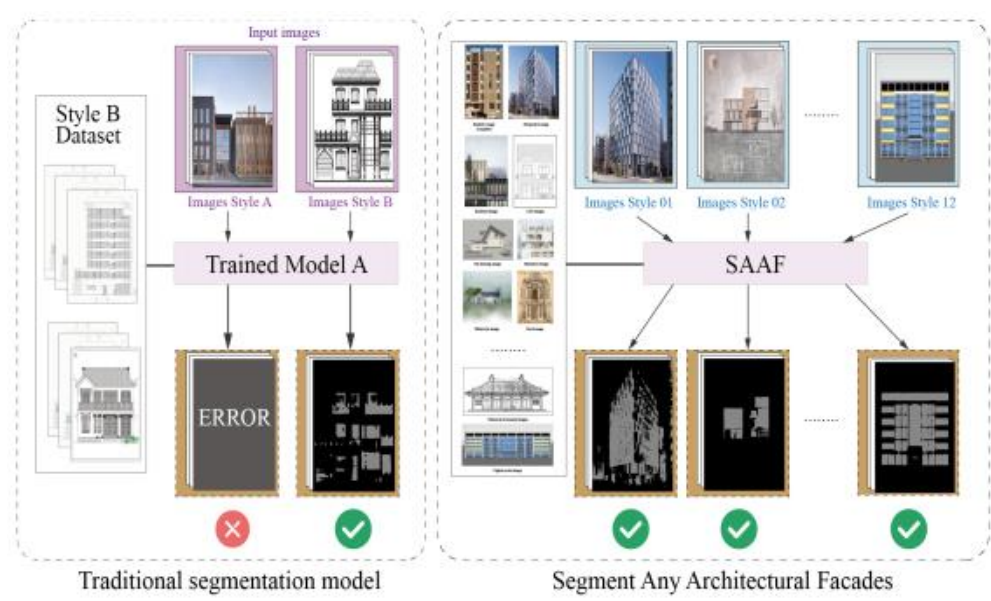

Figure 1
Graphic Abstract

## RELATED WORK

This study proposes a multimodal semantic-guided façade segmentation method by integrating image segmentation, MLLMs, and architectural recognition. It improves accuracy in complex scenes, enhances adaptability under limited annotations via language prompts, and aligns visual features with architectural knowledge, offering a new paradigm for semantic-visual integration in design automation.

## Image segmentation technology

Image segmentation, a core task in computer vision, partitions visual data for better interpretation and processing. Deep learning has advanced end-to-end methods using CNNs and FCNs. For instance, DeepLabV3+ achieves high performance in tasks like COVID-19 lesion segmentation in chest CTs, using multi-scale dilated convolutions and encoder-decoder architecture (Polat, 2022).

However, single-modal image segmentation methods face many challenges in building facade segmentation: (1) Building facade datasets often suffer from limited annotations, blurred boundaries, and stylistic variability, which hinder model generalization. (2) Complex lighting and material reflections hinder the recognition of elements like window frames and decorative details on intricate facades. (3) In addition, Current methods lack sufficient understanding of building semantics and geometry, leading to low accuracy on non-standard or historical structures. Therefore, how to combine multi-modal information and improve the model's adaptability to different lighting conditions and building styles remains an urgent problem to be solved in the field of building facade image segmentation.

## Multimodal large language models

Multimodal Large Language Models (MLLMs) have become a key focus in AI, integrating data from text, images, and point clouds to enhance reasoning (Li et al., 2023a). Models like CLIP, BLIP, and LLaVA enable cross-modal tasks and support perception and decision-making in applications such as autonomous driving (Cui et al., 2024). Their use of language priors is particularly effective for façade segmentation under limited data.

However, current large language models lack domain-specific knowledge of architectural components, limiting their performance in façade segmentation. Moreover, reliance on cross-modal alignment mechanisms can lead to semantic mismatches between images and textual data, reducing recognition stability and accuracy. Future research should focus on enhancing architectural adaptability, improving semantic and geometric understanding of components, and minimizing reliance on large-scale annotated datasets.

## Building façade recognition

Façade recognition, involving automatic detection of elements like walls and windows, is a key task in architectural computer vision. Traditional methods rely on handcrafted features (e.g., SIFT, HOG, Gabor filters; Dalal & Triggs, 2005), using geometry and texture for classification. However, rule-based approaches struggle with generalization across diverse styles, materials, and scales, limiting their practical use.

Recent advances apply deep learning, particularly CNNs, to building façade recognition. For instance, CNNs enable precise segmentation of residential façades for urban renewal (Dai et al., 2021). Researchers are also integrating architectural knowledge into models, using GNNs to capture element relationships and developing domain-specific datasets and metrics.

# DATASET

Traditional datasets usually have problems such as single source, limited style, and uneven scale. Therefore, it is very necessary to construct a high-quality, cross-style building façade dataset. However, due to the diversity of styles and forms among different datasets, different datasets often require processes of collection, processing, and correction, which makes the evaluation of cross-style building façade datasets difficult.

Furthermore, Wall and window segmentation is challenged by complex geometries, dynamic lighting, and occlusions from decorative elements. These factors limit dataset representativeness and reduce model generalization. To address this, we developed a diverse, high-quality façade dataset spanning multiple styles and visual conditions.

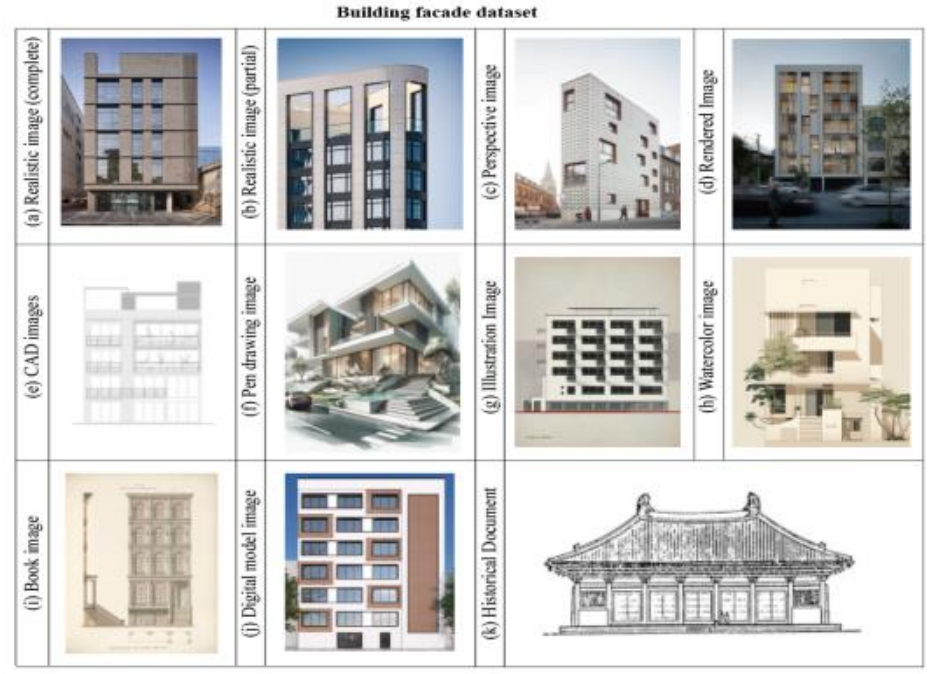

Figure 2 Building facade image dataset: visual examples.

## Building facade dataset

To enhance model generalization, we curated 1,200 high-resolution façade images from platforms like Pinterest®, Behance®, Adobe Stock®, Archdaily®, and Gooood®. As shown in Figure 2, the dataset spans diverse formats — photos, renderings, sketches, CAD drawings, historical records, and stylized façades — capturing a broad range of visual and architectural variation To further improve the generalization ability and stability of the model under different architectural styles and complexity conditions, the dataset is divided into a training set, a test set, and a validation set in a ratio of 7:2:1, and a data deduplication strategy is adopted.

## Referring segmentation dataset

As illustrated in Figure 3, we constructed a dedicated referring segmentation dataset by processing the building façade image corpus and assigning concise textual descriptions focused on wall-window components. The clear description {description} refers to keywords for wall-window segmentation tasks such as daylight-admitting components, glazed sections, transparent surfaces,and fenestration elements. To transform the data into question-answer pairs, we employed a structured prompt format: "User: <Image> Help me segment the objects in this image according to {description}? SAAF: Understood, it is <SEG>."

## Trainable parameters

To preserve the knowledge embedded in the pre-trained multimodal LLMF (specifically, SAAF in this study), we utilized Low-Rank Adaptation (LoRA) to enable efficient fine-tuning while keeping the visual backbone network $E_{\text{enc}}$ entirely frozen to retain its pretrained features. Concurrently, the decoder $D_{\text{dec}}$ was thoroughly fine-tuned to adapt to the downstream task. Additionally, during training, parameters such as embed_tokens, lm_head, and the projection layer Y within the large language model (LLM) were marked as trainable.

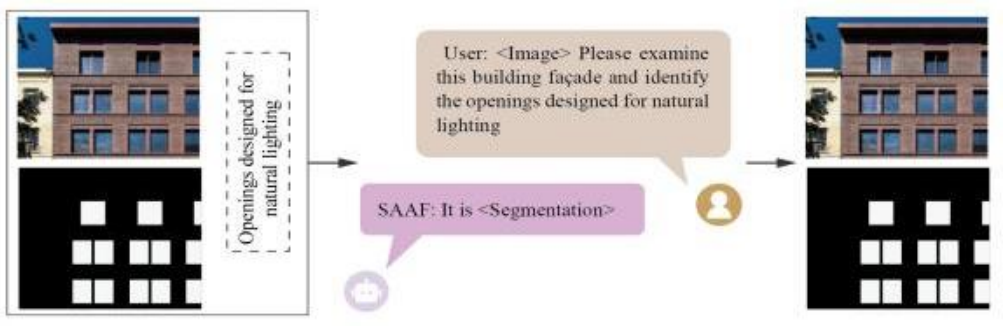

Figure 3 Schematic diagram of the referring segmentation dataset

Notably, this method retains the model’s text and dialogue capabilities, avoiding catastrophic forgetting. This is likely due to: (1) LoRA’s minimal parameter tuning; and (2) VQA data integration, which enhances visual reasoning without harming language output.

## METHODOLOGY

Although models such as Flamingo and BLIP-2 can process multimodal inputs, they do not offer inherent segmentation functionality. VisionLLM attempts to overcome this limitation by embedding polygon-based mask representations within LLMs (Zeng & Dai, 2024). However, this approach introduces significant optimization complexity and exhibits limited generalization when not supported by extensive datasets and computational resources.

To enable façade segmentation within multimodal LLMs, we introduce an embedding-as-mask architecture. A dedicated token, <SEG>, is employed to activate the decoding of its embedding into a detailed high-resolution mask that delineates components such as windows and walls (Figure 4).

Upon receiving a natural language prompt (e.g., "Annotate all windows") denoted as instruction , along with a façade image , the multimodal LLMF jointly processes both inputs and produces a corresponding textual output . This interaction can be formally described as:

$$\hat{\beta}_{txt} = \phi(\alpha_{txt} \oplus \alpha_{img}) \qquad (1)$$

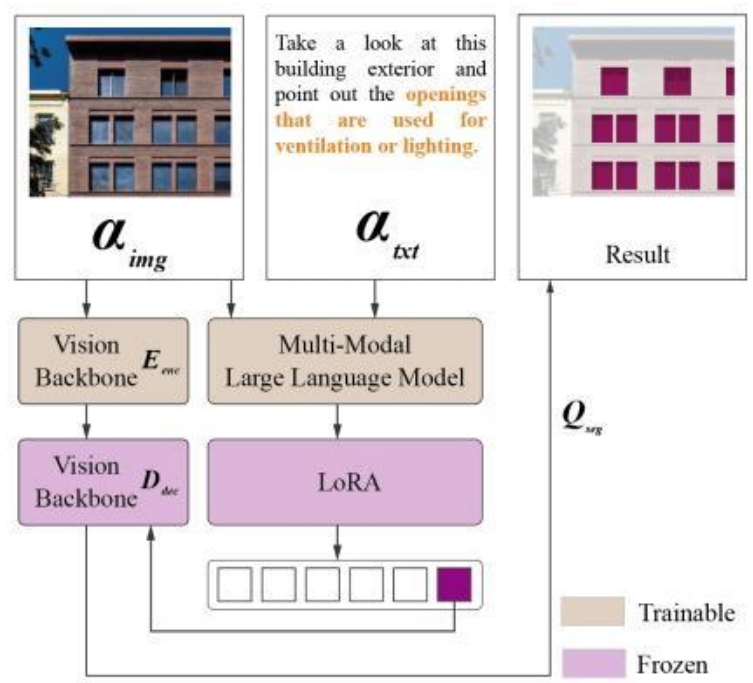

Figure 4 Overview of the SAAF Framework: A Multimodal Segmentation Architecture Integrating Language and Vision Inputs for Building Facade Understanding.

When prompted to generate a binary segmentation mask, the LLM outputs a <SEG> token. Its final-layer embedding $Q_{\mathrm{seg}}$ is projected via an MLP to produce the segmentation embedding. Meanwhile, the vision backbone $E_{\mathrm{enc}}$ extracts visual features $f$ from input $\alpha_{\mathrm{img}}$ . The decoder $D_{\mathrm{dec}}$ then combines $Q_{\mathrm{seg}}$ and $f$ to produce the final segmentation mask. This mechanism supports end-to-end training, enabling the LLM to directly map textual and visual inputs to segmentation masks. It simplifies traditional multi-stage segmentation pipelines and enhances both training efficiency and inference stability. This process can be expressed as:

$$Q_{\mathrm{seg}} = \gamma(\tilde{Q}_{\mathrm{seg}}), f = E_{\mathrm{enc}}(\alpha_{\mathrm{img}}) \qquad (2)$$

$$\mathrm{S} = D_{\mathrm{dec}}(Q_{\mathrm{seg}}, f) \qquad (3)$$

We adopt a joint loss function to optimize model training. By combining the autoregressive cross-entropy loss $L_t$ of text generation and the segmentation mask loss $L_m$, we conduct end-to-end training, enabling the model to simultaneously optimize text understanding ability and the performance of the segmentation task. The total loss $\mathrm{L}$ is a weighted combination of both terms, determined by $\theta_{\mathrm{t}}$ and $\theta_{\mathrm{m}}$ :

$$L = \theta_{\mathrm{t}} L_{\mathrm{t}} + \theta_{\mathrm{m}} L_{\mathrm{m}} \qquad (4)$$

To calculate $L_m$ , we use a combination of per-pixel binary cross-entropy loss and DICE loss, with corresponding loss weights of $\theta_{\mathrm{bce}}$ and $\theta_{\mathrm{dice}}$ , to ensure that the model captures the global mask structure and optimizes boundary details. Given the ground truth target $\alpha_{\mathrm{txt}}$ and the mask, these losses can be expressed as:

$$L_{\mathrm{t}} = CrossEntropy(\hat{\mathrm{y}}_{\mathrm{t}}, \mathrm{y}_{\mathrm{t}}) \qquad (5)$$

$$L_{\mathrm{m}} = \theta_{\mathrm{bce}} \cdot \mathrm{CE_{bin}}(\hat{\mathrm{S}}, \mathrm{S}) + \theta_{\mathrm{dice}} \cdot \mathrm{Dice}(\hat{\mathrm{S}}, \mathrm{S}) \qquad (6)$$

This approach enables multimodal LLMs to perform accurate façade segmentation, extending their function beyond text generation. End-to-end training with embedded integration enhances segmentation accuracy and generalization across varied façade types.

## EXPERIMENT RESULT

Network architecture: We use LLaVA-7B-v1-1 or LLaVA-13B-v1-1 as the basic multimodal LLMF and adopt ViT-H SAM as the visual backbone network Fene. The projection layer $y$ is a multi-layer perceptron (MLP) with [256, 4096, 4096] channels.

Training was performed on a workstation with dual NVIDIA A6000 GPUs (48 GB VRAM each) and an Intel Xeon Gold 6326 CPU, ensuring stable and efficient computation. We trained the model using our custom dataset and the Adam optimizer, with a learning rate of 0.0001, batch size of 24, and up to 100,000 steps to balance convergence and stability. Both text and mask loss weights ($\theta_t, \theta_m$) are set to 0.8. Dice loss ($\theta_{dice}$) and BCE loss ($\theta_{bce}$) are weighted at 0.5 and 2.0, respectively. Training uses a per-device batch size of 2 with gradient accumulation over 10 steps.

## Qualitative segmentation results

We input diverse building facade images into the trained SAAF model, combined with an MLLM, using natural language to guide segmentation. Figure 5 illustrates results across various image types (e.g., real-world photos, renderings, sketches, CAD, and literature drawings), demonstrating strong generalization and accurate wall-window segmentation guided by semantic reasoning.

The proposed model accurately parses window geometry in CAD elevations and reliably distinguishes architectural components. It also maintains high recognition accuracy on real-world building photos under varying lighting, materials, and perspectives. SAAF surpasses conventional models in handling oblique views and perspective distortions, reliably identifying wall-window boundaries with contextual awareness. It also generalizes well to stylized inputs like watercolor renderings and architectural illustrations, demonstrating strong semantic flexibility and multimodal understanding.

In addition, natural language input is introduced during the training process to improve the model’s ability to understand different expression methods. Figure 5 shows that even if the user input does not explicitly contain keywords such as "wall" and "window," SAAF can still accurately extract semantic features and generate correct results.

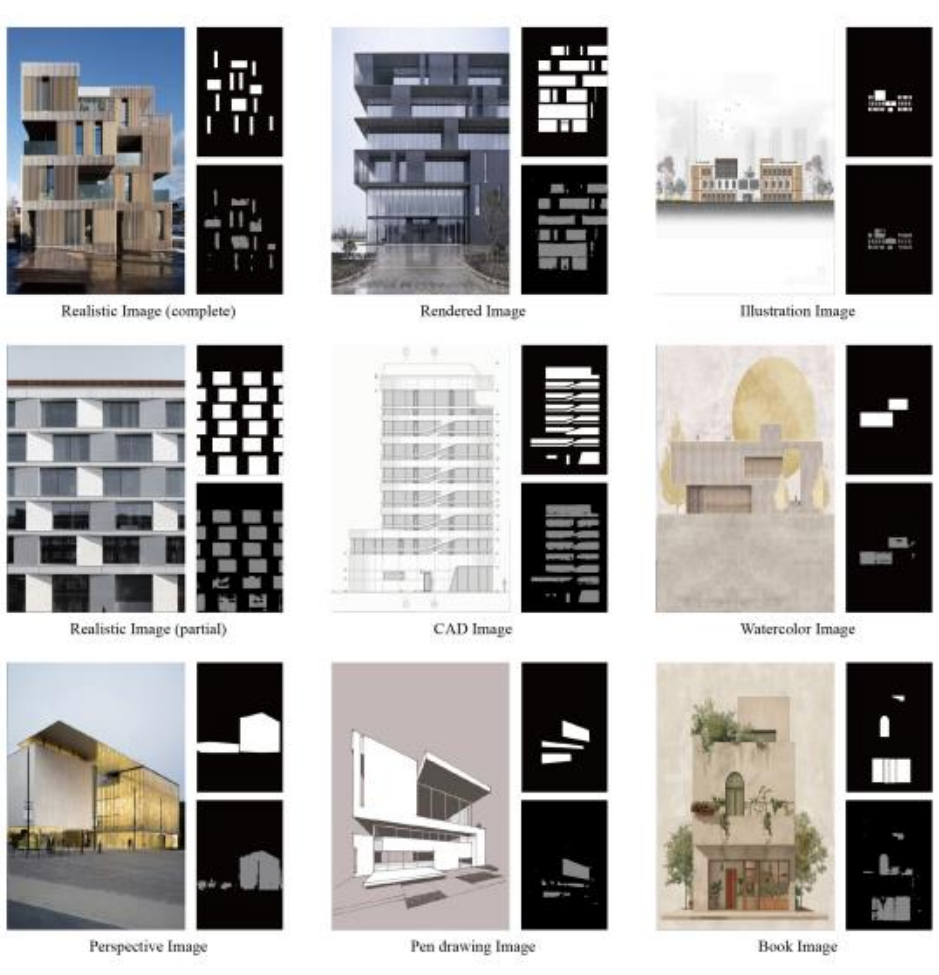

Figure 5 Segmentation results of SAAF on different datasets

Furthermore, as shown in figure 6, SAAF excels not only in basic entity recognition but also in interpreting complex spatial, color, and shape-related language. This highlights its strength in aligning abstract language with image semantics and its potential for cross-modal reasoning and generalization.

To test SAAF’s robustness and generalization, we train on real facade images and validate on a heterogeneous set — including renderings, perspective photos, hand-drawings, and historical images — to simulate real-world style shifts. SAAF is systematically compared with mainstream semantic segmentation models – Fully Convolutional Networks (FCN), U-Net, and High-Resolution Network (HRNet).

As shown in Figure 7, FCN and U-Net struggle with noise, edge loss, and distortion—especially in perspective and line-drawing images. HRNet improves multi-scale fusion but lacks accuracy on sparse or irregular inputs. In contrast, SAAF delivers robust, contour-preserving results across styles, accurately segmenting building components.

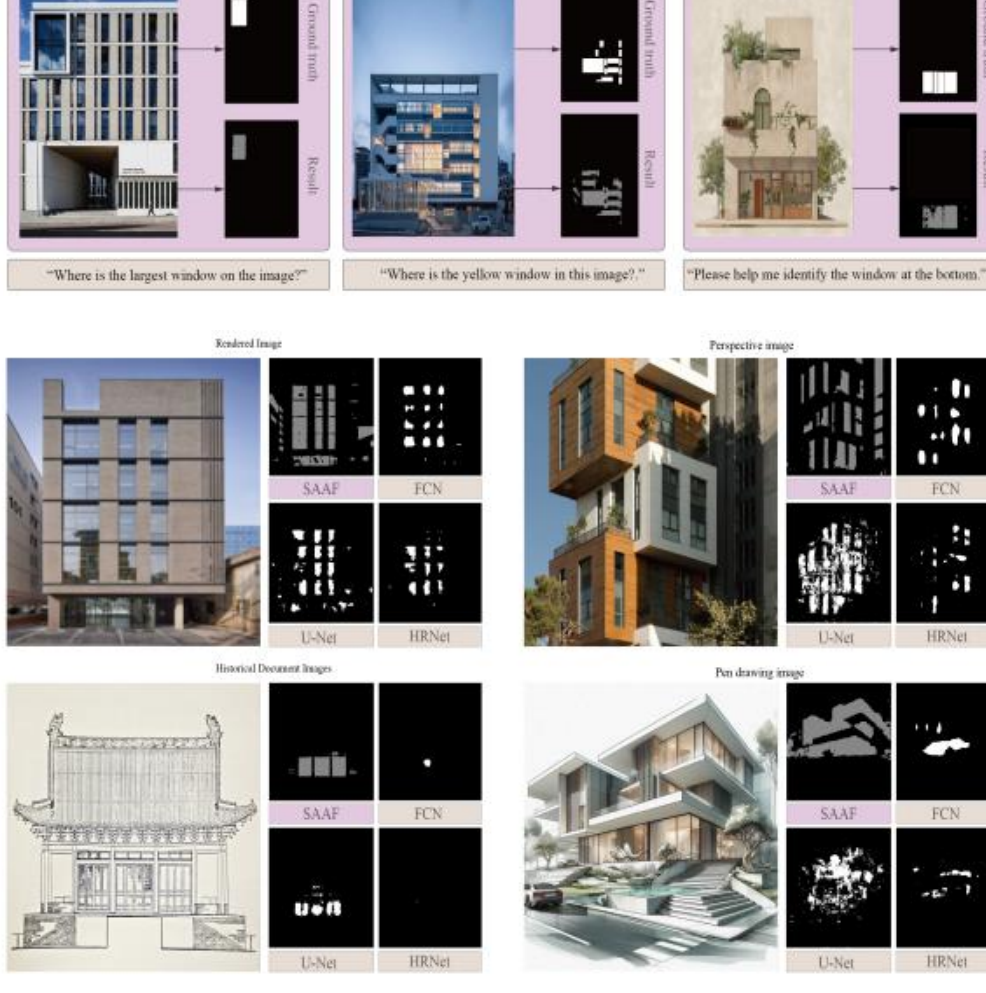

Figure 6
Wall-window segmentation by SAAF based on semantic guidance

Figure 7
Comparison of segmentation effects of different semantic segmentation models on multi-type building images

## Quantitative mIoU evaluation

In the quantitative evaluation, we use mIoU and PA as quantitative evaluation indicators to measure the performance of the model in the segmentation task and conduct a horizontal comparison between SAAF and mainstream semantic segmentation models (see figure 7). SAAF exhibits high mIoU values in various image types (such as pen-and-ink drawings, renderings, and perspective views). Specifically, the mIoU on the pen-and-ink drawing dataset is 0.734, and on real perspective images, it is 0.702, indicating good cross-style adaptability.

Figure 8
mIoU values of different semantic segmentation models on multi-type architectural images

Further comparison results (figure 8) show that the model trained on real-scene images still has good transferability on renderings with similar styles, with mIoU generally higher than 0.5. Among them, U-Net has the best transferability, with mIoU reaching 0.7. However, on images with enhanced style heterogeneity (such as historical archive images), the performance of all models declines as the difference increases, and the mIoU is generally lower than 0.45. Although HRNet has a multi-scale feature fusion structure, it still has difficulty making stable predictions in complex backgrounds.

The experimental results fully demonstrate the efficiency, stability, and wide applicability of the SAAF model in the building facade wall and window segmentation task. It can maintain a high mIoU under various architectural expression methods and real environmental scenarios, and can automatically infer correct segmentation results without explicit rule constraints.

## DISCUSSION

While the proposed SAAF model achieves notable success in automated facade wall-window segmentation, limitations persist. First, datasets under low-resource conditions inadequately cover regional and historical architectural styles, necessitating expanded data scales and the incorporation of self-supervised learning to enhance generalization. Second, the "black-box" nature of deep learning restricts engineering trustworthiness, requiring the integration of interpretable techniques (e.g., attention visualization, decision-path analysis) for transparency. Third, current multimodal fusion strategies underutilize image-text-geometric feature synergies; future work should optimize cross-modal alignment via attention mechanisms or Transformer architectures.

Future directions may include: (1) Extending SAAF’s multimodal framework to medical image segmentation (e.g., organ/lesion recognition) and satellite imagery analysis (geographic parsing, building detection). (2) Develop hybrid systems combining automated segmentation with user fine-tuning via natural language or GUI interactions, thereby enhancing accuracy in complex architectural, BIM, and urban analysis scenarios. (3) Continuously collect diverse facade data (in terms of styles, materials, lighting) and refine models to improve adaptability to unseen conditions.

## CONCLUSION

This study proposes SAAF, a multimodal semantics-guided method for automated wall-window segmentation in architectural facades, integrating deep learning, natural language processing, and multimodal learning to achieve precise segmentation of key components. Validations across multi-source datasets and real-world scenarios demonstrate SAAF's superior segmentation accuracy and automation performance compared to existing methods. While limitations exist, SAAF provides a novel framework with potential applications in building performance optimization, historical structure restoration, and urban digital modeling, driving intelligent transformation in architectural practices.

## AUTHORSHIP INFORMATION

Author Contributions: Conceptualization, P.L.[*], H.Z.[*], and J.Y.[†]; methodology, P.L.[*], H.Z.[*], and J.Y.[†]; software, J.Z. and J.L.; validation, J.L. and M.Z.; formal analysis, J.Z., P.Z., and M.Z.; investigation, P.L.[*] and H.Z.[*]; data curation, J.Z. and P.Z.; writing—original draft preparation, P.L.[*] and H.Z.[*]; writing—review and editing, J.Y.[†]; visualization, J.Z. and H.H.; supervision, J.Y.[†]; discussion, J.Z., H.H., and M.Z. All authors have read and agreed to the published version of the manuscript.

[*]These authors contributed equally to this work.

[†]correspond author.